\documentclass[sigconf]{acmart}

\usepackage{multirow}
\usepackage{multicol}
\usepackage{booktabs}
\usepackage{epsfig}

\usepackage{url}
\usepackage{array}
\usepackage{color}
\usepackage{subfigure}

\graphicspath{ {./images/} }
\setcopyright{acmcopyright}
\copyrightyear{2020}
\acmYear{2020}
\acmDOI{}

\acmConference[Seattle '2020]{ACM International Conference on Multimedia}{}{USA}
\acmBooktitle{Seattle '2020: ACM International Conference on Multimedia}
\acmPrice{}
\acmISBN{}

\begin{document}

\title{%{\normalfont ACM MM Grand Challenge on Human in Events 2020: Track-4}
%\protect\\
Toward Accurate Person-level Action Recognition in Videos of Crowded Scenes
}

\author{Li Yuan}
\authornote{Authors contributed equally to this work. Work done during internship at YITU Technology.}
\email{yuanli@u.nus.edu}
%\orcid{0000-0002-2120-5588}
\author{Yichen Zhou}
%\email{e0251087@u.nus.edu}
\authornotemark[1]
%\email{e0251087@u.nus.edu}
\affiliation{
  \institution{National University of Singapore}
  \institution{YITU Technology}
}

\author{Shuning Chang}
\affiliation{%
  \institution{National University of Singapore}
}

\author{Ziyuan Huang}
\affiliation{%
  \institution{National University of Singapore}
}

\author{Yunpeng Chen}
\affiliation{%
  \institution{YITU Technology}
}
\email{}

\author{Xuecheng Nie}
\affiliation{%
 \institution{YITU Technology}}

\author{Tao Wang}
\affiliation{%
  \institution{National University of Singapore}
}

\author{Jiashi Feng}
\affiliation{%
  \institution{National University of Singapore}}

\author{Shuicheng Yan}
\affiliation{%
  \institution{YITU Technology}}

\renewcommand{\shortauthors}{Yuan and Zhou, et al.}

\begin{abstract}
Detecting and recognizing human action in videos with crowded scenes is a challenging problem due to the complex environment and diversity events. Prior works always fail to deal with this problem in two aspects: (1) lacking utilizing information of the scenes; (2) lacking training data in the crowd and complex scenes. In this paper, we focus on improving spatio-temporal action recognition by fully-utilizing the information of scenes and collecting new data. A top-down strategy is used to overcome the limitations. Specifically, we adopt a strong human detector to detect the spatial location of each frame. We then apply action recognition models to learn the spatio-temporal information from video frames on both the HIE dataset and new data with diverse scenes from the internet, which can improve the generalization ability of our model. Besides, the scenes information is extracted by the semantic segmentation model to assistant the process. As a result, our method achieved an average 26.05 wf\_mAP (ranking 1st place in the 
ACM MM grand challenge 2020: Human in Events).
  
 %This work presents our winning solution to ACM MM 2020 challenge: Large-scale Human-centric Video Analysis in Complex Events; specifically, here we focus on Track4: Person-level Action Recognition in Complex Events~\cite{lin2020human}. Remarkable progress has been made in person-level action recognition in recent years. However, how to localize and recognize the person-level action in crowded and complex scenes has not been well addressed. In this report, we outline the approach in detail for our 1st place solution to the challenge. 
\end{abstract}


\maketitle
\section{Introduction}
As a challenging problem in computer vision, person-level action recognition has been applied in many applications, including human behavior analysis, human-computer interaction, and video surveillance. %Due to the long-term information loss, appearance variance and cluttered background, action recognition is a very challenging task for large scale image and video datasets. 
Recently, significant progress has been made in this area~\cite{slowfast2019ava, aia2020}. However, the person-level action recognition in complex events~\cite{lin2020human} is still relatively new and a challenging problem. In this challenge of person-level action recognition on crowded scenes and complex events, we propose to utilize the scenes information by semantic segmentation model and mine more diversity scenes in data from the internet.

\begin{figure*} [h]
    \centering
    \includegraphics[width=0.9\textwidth]{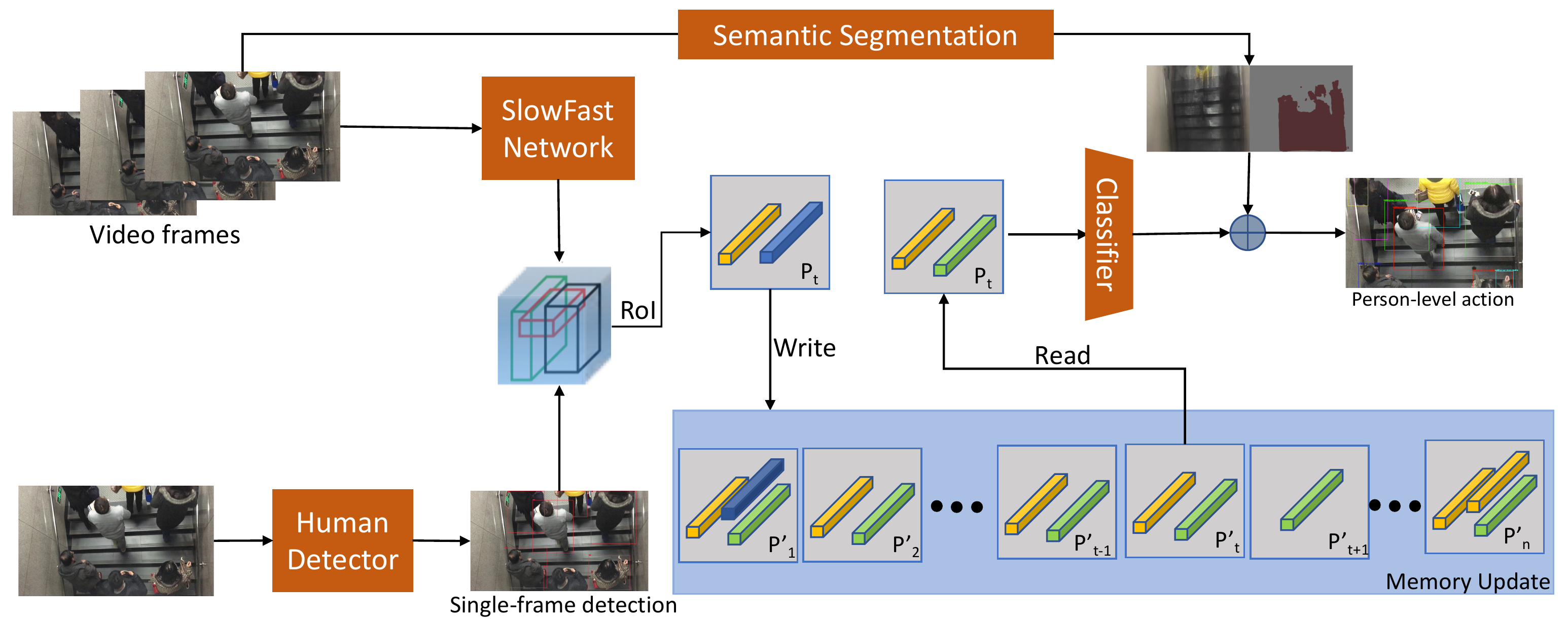}
    \caption{The framework of our person-level action recognition.}
    \label{fig:framework}
\vspace{-5mm}
\end{figure*}

Our overall approach for person-level action recognition in complex events can be divided into two parts, respectively human detection and spatio-temporal action recognition (Figure~\ref{fig:framework}), where the second part is very similar to the methods used by the champion team in the AVA Spatio-temporal Action Localization challenge 2019~\cite{slowfast2019ava}, but we utilize more diversity data and adaptively apply the AVA model to HIE dataset. Firstly, we train a very strong person detector to detect the persons in the video frames. We then build an action detection model using a modified version of the Asynchronous Interaction Aggregation
network~\cite{aia2020}. Finally, we extract the scenes information by semantic segmentation model to boost the performance. 

\section{Human Detection}
The first step of spatio-temporal action localization is to detect the bounding boxes of person. As no validation set in HIE dataset~\cite{lin2020human}, we split the original training set as new training set and validation set. Two splitting strategies are tried: splitting by image frames (5k for validation, 27k for training) and splitting by videos. We found that splitting by image will cause over-fitting and far away from the data distribution of testing set. So we adopt the video-splitting strategy and split video 3,7,8 and 17 as the validation set and the rest videos as train data, in which 5.7k image frames for validation and the reset 27k images frames for training. Based on the train and validation set, we can conduct detection experiments on HIE. All the performance of models is tested by two metrics, Averaged Precision (AP) and MMR~\cite{dollar2011pedestrian}. AP reflects both the precision and recall ratios of the detection results; MMR is the log-average Miss Rate on False Positive Per Image (FPPI) in $[0.01,100]$, is commonly used in pedestrian detection. MR is very sensitive to false positives (FPs), especially FPs with high confidences will significantly harm the MMR ratio. Larger AP and smaller MMR indicates better performance.

\begin{table}[]
\begin{center}
\fontsize{8pt}{12pt}\selectfont
\caption{Performance comparison (AP and mMR) among different detection backbone and methods on HIE dataset. }
\begin{tabular}{l|c|c}
 \toprule
 Methods or Modules & AP (\%) & MMR (\%) \\
 \midrule
 Baseline (ResNet50 + Faster RCNN) & 61.68 & 74.01 \\
 ResNet152 + Faster RCNN &67.32 & 68.17\\
 ResNet152 + Faster RCNN + FPN &69.77 & 64.83\\
 SENet154 +  Faster RCNN + FPN &65.77 & 68.46\\
 ResNeXt101 + Faster RCNN + FPN &69.53 & 63.91\\
 ResNeXt101 + Cascade RCNN + FPN &71.32 & 61.58\\
 ResNet152 +  Cascade RCNN + FPN &71.06 & 62.55\\
 \bottomrule
\end{tabular}
\label{tab:det_methods}
\end{center}
\vspace{-4mm}
\end{table}

\subsection{Detection Frameworks on HIE}
There are mainly two different types of common detection frameworks: one-stage (unified) frameworks~\cite{redmon2016you, redmon2016yolo9000, liu2016ssd} and two-stage (region-based) framework~\cite{girshick2014rich, girshick2015fast, ren2015faster, he2017mask}. Since RCNN~\cite{girshick2014rich} has been proposed, the two-stage detection methods have been widely adopted or modified~\cite{ren2015faster, lin2017focal, lin2017FPN, cai2018cascade, zhou2018object, wang2019few, wang2019distilling}. Normally, the one-stage frameworks can run in real-time but with the cost of a drop in accuracy compared with two-stage frameworks, so we mainly adopt two-stage frameworks on HIE dataset.  

We first investigate the performance of different detection backbone and framework on HIE dataset, including backbone: ResNet152~\cite{he2016deep}, ResNeXt101~\cite{xie2017aggregated} and SeNet154~\cite{hu2018squeeze}, and different framework: Faster-RCNN~\cite{ren2015faster}, Cascade R-CNN~\cite{cai2018cascade}, and Feature-Pyramid Networks (FPN)~\cite{lin2017feature}. The experimental results on different backbone and methods are given in Table~\ref{tab:det_methods}. The baseline model is Faster RCNN with ResNet50, and we search hyper-parameters on the baseline model then apply to the larger backbone. From table~\ref{tab:det_methods}, we can find that the better backbone (ResNet152 and ResNeXt101) and combining advanced methods (Cascade and FPN) can improve the detection performance, but the SENet154 does not get better performance than ResNet152 even it has superior classification performance on ImageNet. So in our final detection solution, we only adopt ResNet152 and ResNeXt101 as the backbone.

\begin{table}[]
\begin{center}
\fontsize{8pt}{12pt}\selectfont
\caption{The effects of using extra data for human detection on HIE dataset.}
\begin{tabular}{l|c|c}
%{ p{3.5cm}||p{1cm}|p{1cm}  }
 \toprule
 Validation set & AP (\%) & MMR (\%) \\
 \midrule
 HIE data & 61.68 & 74.01 \\
 HIE + COCO person &65.83 & 69.75\\
 HIE + CityPerson &63.71 & 67.43\\
 HIE + CrowdHuman &\textbf{78.22} & \textbf{58.33}\\
 HIE + self-collected data &\textbf{69.39} & \textbf{60.82}\\
 HIE + CrowdHuman + COCO + CityPerson &78.53 & 58.63\\
 \textbf{HIE + CrowdHuman + self-collected data}  &\textbf{81.03} & \textbf{55.58}\\
 HIE + all extra data &81.36 & 55.17\\
 \bottomrule
\end{tabular}
\label{tab:det_data}
\end{center}
\vspace{-4mm}
\end{table}

\subsection{Extra Data for HIE}
In the original train data, there are 764k person bounding boxes in 19 videos with 32.9k frames, and the testing set contains 13 videos with 15.1k frames. Considering the limited number of videos and duplicated image frames, the diversity of train data is not enough. And the train data and test data have many different scenes, thus extra data is crucial for training a superior detection model. Here we investigate the effects of different human detection dataset on HIE, including all the person images in COCO (COCO person, 64k images with 262k boxes)~\cite{lin2014microsoft}, CityPerson (2.9k image with 19k boxes)~\cite{zhang2017citypersons}, CrowndHuman (15k images with 339k boxes)~\cite{shao2018crowdhuman} and self-collected data (2k images with 30k boxes). We investigate the effects on different data based on Faster-RCNN with ResNet50 as the backbone. The experimental results are shown in Table~\ref{tab:det_data}. We can find that the CrowdHuman dataset achieves the most improvement compared with other datasets, because the CrowdHuman is the most similar dataset with HIE, and both of the two datasets contain plenty of crowded scenes. COCO person contains two times of images than HIE train data, but merging the COCO person does not bring significant improvement and suffer more than three times train time, thus we only merge HIE with CrowdHuman and self-collected data to take a trade-off between detection performance and train time. 

\begin{table}[]
\begin{center}
\fontsize{8pt}{12pt}\selectfont
\caption{Detection in Crowded Scenes on HIE dataset.}
\begin{tabular}{l|c|c}
 \toprule
 Validation set & AP (\%) & MMR (\%)\\
 \midrule
 ResNet50 + Faster RCNN + extra data & 81.36 & 55.17 \\
 + emd loss  &81.73 & 53.20\\
 + refine module &81.96 & 50.85\\
 + set NMS &82.05 & 49.63\\
 \bottomrule
\end{tabular}
\label{tab:det_crowded}
\end{center}
\vspace{-4mm}
\end{table}

\subsection{Detection in Crowded Scenes}

As there are lots of crowded scenes in HIE2020 dataset, the highly-overlapped instances are hard to detect for the current detection framework. We apply a method aiming to predict instances in crowded scenes~\cite{chu2020detection}, named as ``CrowdDet''. The key idea of CrowdDet is to let each proposal predict a set of correlated instances rather than a single one as the previous detection method. The CrowdDet includes three main contributions for crowded-scenes detection: (1) an EMD loss to minimize the set distance between the two sets of proposals~\cite{tang2014detection}; (3). Set NMS, it will skip normal NMS suppression when two bounding boxes come from the same proposal, which has been proved works in crowded detection; (2). A refine module that takes the combination of predictions and the proposal feature as input, then performs a second round of predicting. We conduct experiments to test the three parts on HIE2020 dataset, and the results are shown in Table~\ref{tab:det_crowded}. Based on the results in the Table, we can find that the three parts do improve the performance in crowded detection. Meanwhile, we apply KD regularization~\cite{yuan2019revisit} in the class's logits of the detection model, which can consistently improve the detection results by 0.5\%-1.4\%.

Finally, based on the above analysis, we train two detection models on HIE by combining extra data with the crowded detection framework: (1). ResNet152 + Cascade RCNN + extra data + emd loss + refine module + set NMS + KD regularization, whose AP is 83.21; (2). ResNeXt101 + Cascade RCNN + extra data + emd loss + refine module + set NMS + KD regularization, whose AP is 83.78;  Then two models are fused with weights 1:1.

%\begin{figure*} [h]
%    \centering
%    \includegraphics[width=1.0\textwidth]{25_rgb.jpg}
%   \caption{Scene parsing visualization for video 25}
%    \label{fig:25_rgb}
%\end{figure*}

%\begin{figure*} [h]
%    \centering
%    \includegraphics[width=1.0\textwidth]{30_rgb.jpg}
%    \caption{Scene parsing visualization for video 30}
%    \label{fig:30_rgb}
%\end{figure*}

\section{Person-level Action Recognition}
The HIE ~\cite{hie2020} training set contains 33559 human action instances from 14 action categories. The actions are annotated for all individuals in every 20 frames, we call the frames which contain action annotations as the \textbf{key frames}. As mentioned in ~\cite{hie2020}, the distribution of all action classes is extremely unbalanced. Some action classes such as running-alone, running-together, sitting-talking, fighting and fall-over each have less than 400 action instances, while there are 11.7k actions labeled as walking-alone. Moreover, the spatial sizes of the persons in HIE only occupy less than 2\% of the area of the whole frame on average. Therefore it is an extremely challenging task to perform spatio-temporal action localization on this dataset. To mitigate the problem of lack of data, we collected some extra data as mentioned previously and annotated around 2.9k action instances on them.

Next, we will describe the methods we used for the person-level action recognition task. For every key frame, We first sample frames temporally-centered around the key frame to form a short trimmed video clip. The trimmed clip is then fed into the backbone video feature extractor model which outputs feature maps representing the video clip. We then apply RoIAlign~\cite{he2017mask} along the spatial dimensions, and global average pooling along the temporal dimension following ~\cite{feichtenhofer2019slowfast} to extract features for each person. The person features of the current clip together with the features in the neighbor clips are then combined for a few interaction modules for the final action prediction.

\subsection{Action detection Model}
We employ the Asynchronous Interaction Aggregation (AIA) network~\cite{aia2020} as the main action detection model for this task. The backbone of the model is SlowFast 8×8 ResNet-101 (SlowFast8x8-R101) ~\cite{feichtenhofer2019slowfast} which is pretrained on the Kinetics-700 dataset~\cite{k700}. The whole AIA model (with dense serial Interaction Aggregation) is then trained on the AVA dataset~\cite{gu2018ava} which is a large-scale spatio-temporal action localization dataset. We take the AVA trained model open-sourced by the authors of AIA~\cite{aia2020} as our base model. As the actions in HIE are all human-centric and have no interactions with other objects, unlike the actions in AVA, we remove the person-object interactions modules in the AIA. Moreover, as the duration for each video in HIE is much shorter than AVA, and the movement of the persons are generally much quicker in HIE, we reduce the number of neighbor clip $L$ to 1 for the temporal interactions in AIA.

\subsection{Dataset split}
We tried splitting the dataset by videos to train and validation set similarly to the detection training. However, several action labels would have very few positive instances in the training and/or validation set, which makes it hard to evaluate performance using this kind of split strategy. Therefore, we performed a few rounds of initial experiments with the above train-validation split to verify the correctness of our implementation. After that, we switch to training with the full training set without validation, and perform testing on the official testing set given.

\subsection{Training Details}
For each trimmed clip around a key frame, we sample 32 frames and 8 frames within the clip for the slow and fast branch in the feature extractor respectively. For the spatial dimensions, during training, we resize the shorter side of the frames to a length randomly sampled from (320, 352, 384, 416, 448), and pad the longer side to 2.5 times of the shorter side. We apply bounding box jittering of 10\% and random horizontal flipping to improve the generalization ability. We use a dropout rate of 0.2 before the classification prediction, and employ the binary cross-entropy loss on sigmoid activated logits as supervision during training. Training is performed with SGD optimizer for a total of 160k iterations with BatchNorm~\cite{ioffe2015batch} statistics being frozen. We train each model on 8 GPUs with 2 clips per GPU, which makes the total batch-size 16. The initial learning rate is 0.01, and is reduced by a factor of 10 after 96k and 128k iterations. We warm-up the learning rate linearly for the first 2k iterations following~\cite{aia2020}. 

\subsection{Weighted sampling}
Due the lack of training samples for some classes, we observed that the score distribution for the rare classes is near zero after the training. To mitigate the imbalance problem among the action classes, we perform weighted clip sampling during training where we assign higher weights to clips that contain rare actions (such as sitting, running, etc) and lower weights to clips that contain frequent actions (such as walking and standing). After applying the weighted sampling strategy, the ratio of the number of occurrence of maximum sampled action to the minimum sampled action is reduced from around 120 to around 10 during training.

\begin{figure}[t!]
\begin{center}
\end{center}
\subfigure{
\includegraphics[scale=0.0535]{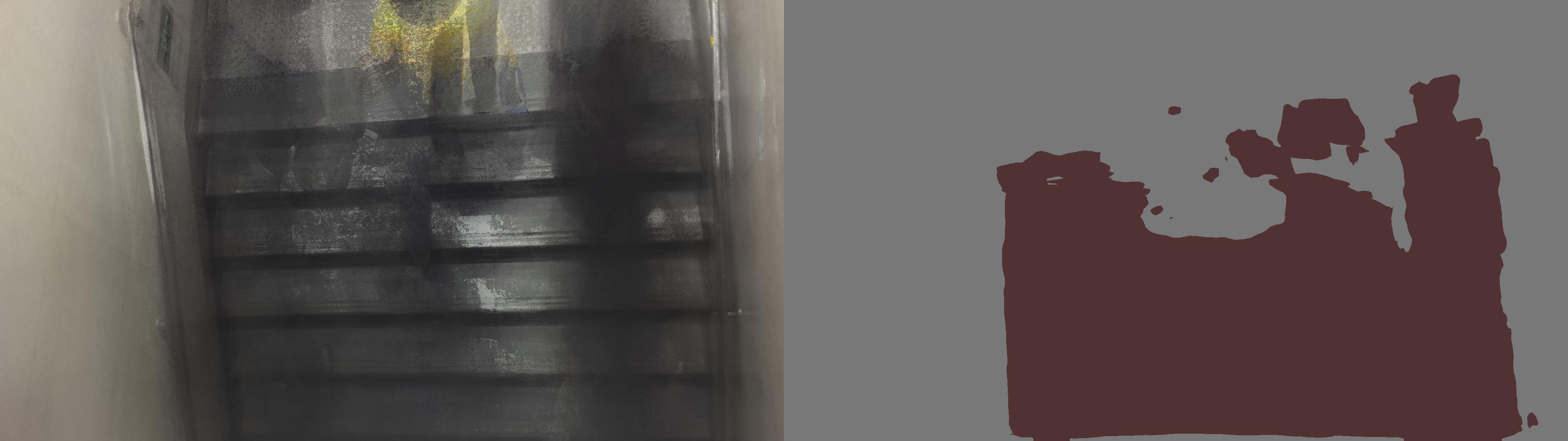}}
\subfigure{
\includegraphics[scale=0.16]{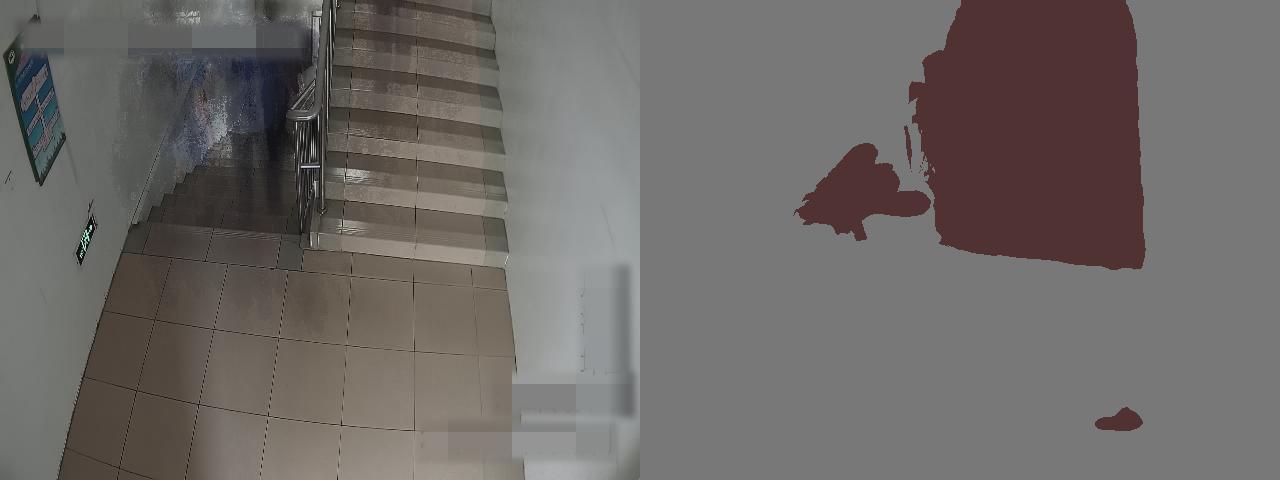}}
\caption{Scene parsing visualization for video 25 and 30. }
\label{fig:visualization}
\vspace{-6mm}
\end{figure}

\subsection{Inference Details}
For inference, the process is similar to that of training, except that instead of ground-truth person boxes, we used the predicted person boxes to perform RoIAlign after the feature extraction. We performed multi-scale and flip testing with 5 scales at (320, 352, 384, 416, 448), and average the predictions scores as the prediction output by the model. For the final submission, we made we only used one single AIA (SlowFast8x8-R101) model without a multi-model ensemble.  

We noticed that certain actions such as walking-up-down-stairs is closely related to staircases in the scene, as they are supposed to be. We can confidently say that the action label of a person is walking-up-down-stairs when the person is above the staircase. Therefore, we use a semantic segmentation model trained on the ADE20K~\cite{zhou2016semantic, zhou2017scene} dataset to perform scene parsing on the videos, so that we can re-weight the scores depending on the location of the person bounding box in the scene. Figure ~\ref{fig:visualization} are the visualizations of scene parsing of hm\_in\_passage (ID:25) and hm\_in\_stair (ID:30) respectively. 

Moreover, certain similar actions such as walking-together, running-together and gathering are group actions that require other people to be physically around. Therefore depending on the maximum intersection-over-union (IoU) among the neighbor person boxes to the target person bounding box, we re-weight the score from 'together' actions to the corresponding 'alone' actions if there is no other person around the target person.

\section{Person-level Action Recognition Results}
Our final submission achieved 26.05\% wf-mAP@avg on the official testing set, which ranked 1st in the competition. The detailed results for each video and the overall performance of the top two teams are shown in table \ref{tab:act_result}. As we can see we perform much better on a few videos such as hm\_in\_bus, hm\_in\_subway\_station, hm\_in\_fighting4 and hm\_in\_restaurant. However, there are also several videos that we did not perform as well as the 2nd place submission, such as the hm\_in\_dining\_room2 video. We suspect that this is because our person detection model is able to detect persons even when they are occluded, but it is hard for the downstream action model to classify the actions correctly, hence hurting the overall performance. % \yp{Highlight the best accuracy in different color}

\begin{table} [t]
\fontsize{8pt}{12pt}\selectfont
\begin{center}
\caption{The top-2 results of HIE2020 testing set. The evaluation metric is wf-mAP@avg(\%).}
\label{tab:act_result}
% \small
\begin{tabular}{l|r|r}
\toprule
Video Name & 1st Place (Ours) & 2nd Place (VM) \\
\midrule
hm\_in\_waiting\_hall    & 27.57  & 23.90 \\
hm\_in\_bus              & 49.30  & 4.73  \\
hm\_in\_dining\_room2    & 12.23  & 16.71 \\
hm\_in\_lab2             & 36.27  & 37.87 \\
hm\_in\_subway\_station  & 28.89  & 10.07 \\
hm\_in\_passage          & 29.77  & 22.20 \\
hm\_in\_fighting4        & 44.10  & 22.08 \\
hm\_in\_shopping\_mall3  & 31.36  & 41.08 \\
hm\_in\_restaurant       & 20.61  & 10.18 \\
hm\_in\_accident         & 32.78  & 27.99 \\
hm\_in\_stair3           & 29.87  & 30.03 \\
hm\_in\_crossroad        & 57.92  & 49.08 \\
hm\_in\_robbery          & 17.36  & 11.79 \\
\midrule
Overall & \textbf{26.05} & 25.48 \\
\bottomrule
\end{tabular}
\end{center}
\vspace{-5mm}
\end{table}

\section{Conclusion}
In this paper, we describe the approach we used in the Person-level Action Recognition in Complex Events track in the HIE2020 Challenge, which achieved 1st place on the testing set. The overall framework is similar to other state-of-the-art spatio-temporal action localization methods where we first detect the persons in some key frames, and then perform action classification using with the help of 3D RoIAlign on the features extracted on a trimmed video clip. We perform extensive experiments and tuning for our person detector which helps to reduce the number of possible false negatives due to missing to localize a person. Then we employ a state-of-the-art action detection model and trained with extra self-collected data in addition to the whole training set. Furthermore, we add some post-processing operations according to human understanding of the actions. 

\bibliographystyle{ACM-Reference-Format}
\bibliography{HIE}

\end{document}